# Super-Resolved Image Perceptual Quality Improvement via Multi-Feature Discriminators

Xuan Zhu[a,*], Yue Cheng[a], Jinye Peng[a], Rongzhi Wang[a], Mingnan Le[a], Xin Liu[a]

[a]Shool of Information Science and Technology of Northwest University, Xi'an, People's Republic of China

**Abstract.** Generative adversarial network (GAN) for image super-resolution (SR) has attracted enormous interests in recent years. However, the GAN-based SR methods only use image discriminator to distinguish SR images and high-resolution (HR) images. Image discriminator fails to discriminate images accurately since image features cannot be fully expressed. In this paper, we design a new GAN-based SR framework GAN-IMC which includes generator, image discriminator, morphological component discriminator and color discriminator. The combination of multiple feature discriminators improves the accuracy of image discrimination. Adversarial training between the generator and multi-feature discriminators forces SR images to converge with HR images in terms of data and features distribution. Moreover, in some cases, feature enhancement of salient regions is also worth considering. GAN-IMC is further optimized by weighted content loss (GAN-IMCW), which effectively restores and enhances salient regions in SR images. The effectiveness and robustness of our method are confirmed by extensive experiments on public datasets. Compared with state-of-the-art methods, the proposed method not only achieves competitive Perceptual Index (PI) and Natural Image Quality Evaluator (NIQE) values but also obtains pleasant visual perception in image edge, texture, color and salient regions.

**Keywords.** image super-resolution, multi-feature discriminators, perceptual quality, weighted content loss.

*Xuan Zhu, E-mail: xuan_zhu@126.com).

## 1 Introduction

Super-resolution (SR) technique is the task of estimating the high-resolution (HR) image from one or a sequence of low-resolution (LR) observations at a lower cost. The improvement of image resolution contributes to accurately recognizing and understanding the image. It has an urgent requirement for SR in many fields, such as computer vision, medical image processing and remote sensing image processing[1]. SR methods can be mainly divided into three categories: interpolation-based methods[3], reconstruction-based methods and learning-based methods. Learning-based methods can be divided into shallow learning SR methods and deep learning SR methods. The shallow learning methods, such as neighbor embedding methods[4], sparse coding methods[6,6] and anchored neighborhood regression methods[5], learn the nonlinear mapping relationship between HR and LR image patches. Deep learning-based methods directly learn end-to-end mapping function between HR and LR images, which is represented by the parameters of convolutional neural networks (CNNs). Recently, CNN for SR has made remarkable improvements.

In 2015, Dong et al.[9] first applied a lightweight CNN to super-resolve an LR image, and the SR network was optimized by minimizing the pixel loss function. It demonstrated excellent reconstruction quality compared with state-of-the-art methods at that time due to the close correlation between pixel loss function and Power Signal-to-Noise Ratio (PSNR). This research is a milestone for SR. Subsequently, considerable researches[10,13,14,15,16] that minimized pixel loss function to train CNNs had been conducted and PSNR values had been dramatically improved. However, the studies[17-19] pointed out that SR results with good visual quality reflected by PSNR values were inconsistent with or even contrary to the subjective evaluation of human observers. Blurry edges and over-smooth textures were shown in SR results while having a high PSNR value. Both Perceptual Index (PI)[25] and Natural Image Quality Evaluator (NIQE)[28] are brought up to evaluate SR results in terms of perceptual quality.

In order to improve SR images visual quality, researchers have introduced different loss functions to optimize SR networks. In 2017, Ledge et al.[17] presented a generative adversarial network (GAN)[8] composed of a generator and an image discriminator for SR. The generator is used to generate SR results. The discriminator is used to determine SR images and HR images. Adversarial learning between generator and discriminator encourages some or several types of data of SR images to be similar to that of HR images. Park et al.[22] proposed a feature discriminator that distinguishes SR image from HR image by feature maps to produce high-frequency details.

Considering that morphological component and color are highly correlated with image visual quality, a natural idea is to introduce morphological component discriminator and color discriminator to identify images. In this paper, we design a new GAN-based SR framework GAN-IMC composed of a generator and multi-feature discriminators. Multi-feature discriminators consist of image discriminator, morphological component discriminator and color discriminator. Image discriminator, like in the standard GAN[17,19], discriminates images by the pixel value. Morphological component discriminator discriminates images by the edge and texture information of images. Color discriminator discriminates images by the color of images. Multi-feature discriminators in GAN-IMC ensure that the edge and texture of SR results are enhanced, and the color misalignment is avoided. SR results generated by GAN-IMC are more consistent with the HR images in pixel, edge, texture and color, and image visual quality is improved.

Moreover, it is well known that visual attention is drawn to salient regions where multiple features are aggregated. Therefore, we propose a weighted content loss function based on human visual attention mechanism to expand the difference between salient and non-salient regions in the image. GAN-IMCW can be obtained by introducing weighted content loss to optimize GAN-IMC. GAN-IMCW significantly improves visual perceptual quality of salient regions in SR results.

Our main contributions are as follows:
1. We design a new SR framework that has multi-feature discriminators to improve image visual perceptual quality. Adversarial learning in terms of morphological component and color contributes to producing pleasant SR results.
2. We propose a weighted content loss function based on human visual attention mechanism. The feature-rich regions in SR results are highlighted.

3. A large number of experimental results show that the proposed method achieves significant improvement on the image perceptual quality assessment metrics (PI and NIQE).

This paper is organized as follows. Sec. 2 briefly reviews the development of SR and image quality evaluation. We describe the proposed SR framework and elaborate on the training procedure in Sec. 3. The experimental results and their evaluation are shown in Sec. 4. Finally, we conclude our work in Sec. 5.

## 2 Related Work

*2.1 Super-Resolution*

The disagreement between objective evaluation results and human subjective observations leads to two research directions: PSNR-driven SR and perceptual-quality driven SR[2].

**PSNR-driven SR.** The PSNR-driven SR methods optimize SR networks by minimizing pixel loss function. Kim et al.[10,13] exploited recursive learning and residual learning[11] to build deeper networks to improve the PSNR values of SR results. Zhang et al.[12]12 applied a residual channel attention network that rescales channel-wise features to learn high-frequency information. Tong et al.[15] combined different level feature maps using dense skip connections to boost model performance. Haris et al.[24] exploited iterative up- and down-sampling layers to feedback projection errors and concatenated feature maps across up- and down-sampling stages to super-resolve image for large scaling factors ($\times$ 8). All these methods aim to improve PSNR value while often producing visually unpleasing SR results.

**Perceptual-quality driven SR methods.** The perceptual-quality driven SR methods introduce different loss functions to optimize the SR network for improving visual quality. Johnson et al.[20] proposed content loss that measures Euclidean distance between feature maps of SR images and HR images. Roey et al.[23] proposed contextual loss that measures cosine distance between the feature maps of images. Ledge et al.[17] applied adversarial loss to optimize SR network. Cheon et al.[26] proposed perceptual image content loss that measures the difference between images after applying Discrete Cosine Transform (DCT) and differential operation on SR images and HR images. Sajjadi et al.[19] used texture loss[21] to ensure the consistent style between images. A combination of multiple loss functions has widely been used to produce visually satisfactory SR results.

*2.2 Perceptual Image Quality Evaluation*

For SR, the classic objective evaluation methods (e.g., PSNR) evaluate the image by statistically measuring distortion values between the SR image and the HR image, and mainly concern the difference between the pixel values of the same position of the two images. To evaluate SR images perceptual quality accurately, perceptual index (PI) was proposed in the PIRM Challenge on perceptual SR[25], defined as follows:

$$PI = \tfrac{1}{2}\big((10 - Ma(I)) + NIQE(I)\big) \quad (1)$$

where $I$ is the evaluated image, $Ma(\cdot)$27 is the no-reference image quality evaluation function[27] and $NIQE(\cdot)$ is the quality evaluator score function[28]. Ref. 25 discussed the correlation between image quality assessment and human-opinion-scores. Both NIQE and PI have a high correlation with human-opinion-scores. The lower PI value and NIQE value denote the better perceptual quality.

In this study, we use NIQE and PI indicators to evaluate SR images visual quality.

## 3  GAN-IMC

The HR image $I_{HR}$ is degraded into the LR image $I_{LR}$. The degradation process is defined as follows:

$$I_{LR} = (I_{HR} \otimes k) \downarrow_s + n \qquad (2)$$

where $k$ denotes blur operation, $\downarrow$ denotes down-sampling operation, s denotes scaling factor, and $n$ denotes noise addition operation. The SR network is used to predict the lost high-frequency information during degradation. The SR implementation based on convolution neural network is described as follows:

$$I_{SR} = G(I_{LR}) \qquad (3)$$

where $G$ denotes SR network that takes $I_{LR}$ as input, and outputs the SR image $I_{SR}$.

Our GAN-IMC architecture is composed of generator network $G$ and multi-feature discriminator network $D$, as shown in Fig. 1. Multi-feature discriminator network include image discriminator $D_{img}$, morphological component discriminator $D_{mc}$ and color discriminator $D_c$. GAN-IMC aims to make SR results successfully deceive the discriminators $D_{img}$, $D_{mc}$ and $D_c$, and SR results are similar to HR images in data distribution.

GAN-IMC is obtained by alternately optimizing discriminators $D_{img}$, $D_{mc}$, $D_c$ and generator $G$. Multi-feature discriminator network training is given in Sec. 3.1. The parameters of the generator network are updated by optimizing the perceptual loss function, which is described in Sec. 3.2.

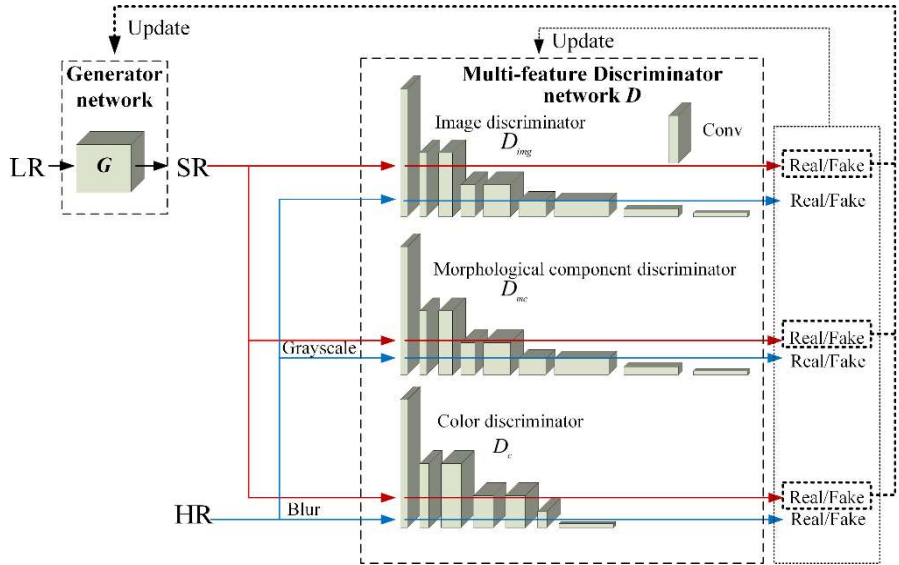

**Fig. 1** The network architecture of GAN-IMC.

## 3.1 Multi-Feature Discriminator Network Training

Multi-feature discriminators $D_{img}$, $D_{mc}$ and $D_c$ are trained to distinguish SR images from HR images in terms of pixel, edge, texture and color, respectively. The combination of multiple discriminators avoids the limitation on simple use of $D_{img}$ and improves the accuracy of discrimination.

**Image discriminator $D_{img}$.** Image discriminator takes the images as input and outputs the probability of the input image being the HR image. The architecture of $D_{img}$ is shown in Table 1(a).

**Morphological component discriminator $D_{mc}$.** Human vision is sensitive to the morphological component of images, which mainly contains edge and texture information. $D_{mc}$ is built to discriminate images by the edge and texture information. Gray image without color and brightness can better highlight the edge and texture. We take the gray image as input of $D_{mc}$. The gray image is obtained:

$$I^g = 0.299 \cdot I_r + 0.578 \cdot I_g + 0.114 \cdot I_b \tag{4}$$

where $I^g$ is the gray image. $I_r$, $I_g$ and $I_b$ denote red, green and blue components of the input image, respectively. To train $D_{mc}$, we minimize the loss function $L_{D_{mc}}$ as follows:

$$L_{D_{mc}} = -\log(D_{mc}(I^g_{HR})) - \log\left(1 - D_{mc}(I^g_{SR})\right) \tag{5}$$

where $I^g_{SR}$ and $I^g_{HR}$ are gray SR image and gray HR image. $D_{mc}(\cdot)$ denotes the probability that the morphological component of the input image belongs to the HR image. The architecture of $D_{mc}$ is shown in Table 1(b).

**Table 1** The architecture of multi-feature discriminators.

| (a) Image discriminator | (b) Morphological component discriminator | (c) Color discriminator |
|---|---|---|
| Conv (k3, s1, n64), LeakyRelu | Conv (k3, s1, n64), LeakyRelu | Conv (k11, s4, n48), LeakyRelu |
| Conv (k3, s2, n64), LeakyRelu, BN | Conv (k3, s2, n64), LeakyRelu, BN | Conv (k5, s2, n64), LeakyRelu, BN |
| Conv (k3, s1, n128), LeakyRelu, BN | Conv (k3, s1, n128), LeakyRelu, BN | Conv (k3, s1, n128), LeakyRelu, BN |
| Conv (k3, s2, n128), LeakyRelu, BN | Conv (k3, s2, n128), LeakyRelu, BN | Conv (k3, s2, n128), LeakyRelu, BN |
| Conv (k3, s1, n256), LeakyRelu, BN | Conv (k3, s1, n256), LeakyRelu, BN | Conv (k3, s1, n128), LeakyRelu, BN |
| Conv (k3, s2, n256), LeakyRelu, BN | Conv (k3, s2, n256), LeakyRelu, BN | Conv (k3, s2, n64), LeakyRelu, BN |
| Conv (k3, s1, n512), LeakyRelu, BN | Conv (k3, s1, n512), LeakyRelu, BN | FC 1024 |
| Conv (k3, s1, n512), LeakyRelu, BN | Conv (k3, s1, n512), LeakyRelu, BN | FC 1 |
| FC 1024 | FC 1024 | |
| FC 1 | FC 1 | |

$n$, $k$ and $s$ denote the number of Conv filters, the size of filters and the stride, separately.

Both $D_{img}$ and $D_{mc}$ have deep network architecture which apply deep semantic feature maps to distinguish $I_{SR}$ and $I_{HR}$.

**Color discriminator $D_c$.** Visual system is also sensitive to the main color, brightness and contrast of objects in natural images. Therefore, we apply a Gaussian blur kernel to blur natural image for reserving main color, brightness and contrast. The color discriminator $D_c$ takes blurred image as input and discriminates images by color, brightness and contrast. The blurred image is obtained via blurred convolution $B$:

$$I^B = I * B \tag{6}$$

where $I$ is the input image, $I^B$ is the blurred image, $*$ denotes the convolution operation, and $B$ is the blur filter. The size of $B$ is $21 \times 21$, the stride is 1, and the weights are fitted to Gaussian distribution, easily calculated as follows:

$$B(x,y) = \frac{1}{2\pi\sigma^2} \exp\left(-\frac{(x-\mu_x)^2}{2\sigma_x^2} - \frac{(y-\mu_y)^2}{2\sigma_y^2}\right) \tag{7}$$

where $x$ and $y$ are horizontal and vertical indexes of blur filter, $\mu_{x,y}$ and $\sigma_{x,y}$ are the mean and variance of $x$ and $y$, separately. We set $\mu_{x,y} = 0$ and $\sigma_{x,y} = \sqrt{3}$. To train color discriminator, we minimize the loss function $L_{D_c}$ as follows:

$$L_{D_c} = -\log(D_c(I_{HR}^B)) - \log(1 - D_c(I_{SR}^B)) \tag{8}$$

where $I_{SR}^B$ and $I_{HR}^B$ are blurred SR image and blurred HR image. $D_c(\cdot)$ denotes the probability that the color of input image belongs to HR image. The architecture of $D_C$ is shown in Table I(c).

*3.2    Generator Network Training*

The architecture of our generator network $G$ is derived from Ref. 18, but the loss function is different from it. Our generator network $G$ is trained with perceptual loss function composed of the pixel loss $L_{pixel}$, adversarial loss $L_{adv}$ and weighted content loss $L_{wc}$, as shown in Fig. 2.

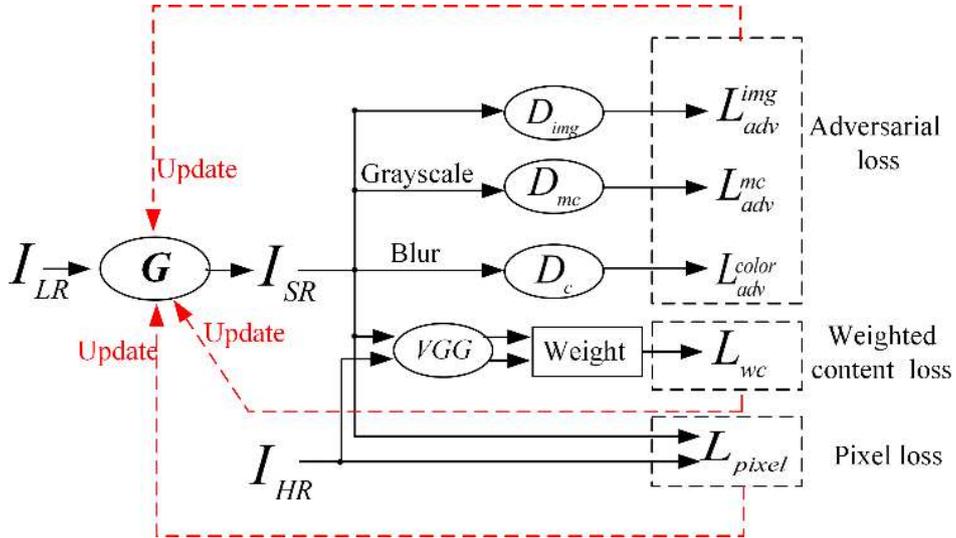

**Fig. 2** The generator network training.

### 3.2.1 Pixel Loss Function

Pixel loss function $L_{pixel}$ constrains the SR image $I_{SR}$ to be close enough to the HR image $I_{HR}$ on the pixel values. We measure Mean Square Error (MSE) between $I_{SR}$ and $I_{HR}$, and $L_{pixel}$ is defined as follows:

$$L_{pixel} = \frac{1}{hwc}\sum(I_{SR} - I_{HR})^2 \qquad (9)$$

where $h$, $w$ and $c$ are the height, width and number of channels of the SR image, respectively.

### 3.2.2 Adversarial Loss Function

Our adversarial loss function is composed of image adversarial loss $L_{adv}^{img}$, morphological component adversarial loss $L_{adv}^{mc}$ and color adversarial loss $L_{adv}^{color}$. Minimizing adversarial loss function makes generator $G$ learn to create solutions that are highly similar to HR images in terms of image, edge, texture and color.

**Image adversarial loss.** The image adversarial loss $L_{adv}^{img}$ is defined as follows:

$$L_{adv}^{img} = -\sum \log D_{img}(I_{SR}) \qquad (10)$$

where $D_{img}(I_{SR})$ is the output of the discriminator $D_{img}$ when $I_{SR}$ is taken as input.

**Morphological component adversarial loss.** The morphological component adversarial loss $L_{adv}^{mc}$ is defined as follows:

$$L_{adv}^{mc} = -\sum \log D_{mc}(I_{SR}^g) \qquad (11)$$

where $D_{mc}(I_{SR}^g)$ is the output of the discriminator $D_{mc}$ when the morphological component of $I_{SR}$ is taken as input.

**Color adversarial loss.** The color adversarial loss $L_{adv}^{color}$ is defined as follows:

$$L_{adv}^{color} = -\sum \log D_c(I_{SR}^B) \qquad (12)$$

where $D_c(I_{SR}^B)$ is the output of the discriminator $D_c$ when the color of $I_{SR}$ is taken as input.

The total adversarial loss function $L_{adv}$ is calculated as follows:

$$L_{adv} = L_{adv}^{img} + 10^{-1}L_{adv}^{mc} + 4 \times 10^{-3}L_{adv}^{color} \qquad (13)$$

### 3.2.3 Weighted Content Loss Function

The studies[17,19,20] have pro ved that the introduction of content loss function improves the visual quality of SR results. In Ref. 22, the low-level feature maps and high-level feature maps of the image are extracted through the $\phi_{2,2}$ and $\phi_{5,4}$ of pre-trained VGG-19 network*, respectively.

We propose a modified weighted content loss function which combines weighted low-level features with high-level semantic features. Minimize the Euclidean distance between two weighted low-level feature maps from $I_{SR}$ and $I_{HR}$ enhances the features of salient regions while weakening the features of non-salient regions. The high-level feature maps are also applied to constrain the semantics of the whole image. The weighted content loss function $L_{wc}$ is formulated as follows:

$$L_{wc} = L_{low-level} + 10^{-5} L_{high-level} \tag{14}$$

$$L_{low-level} = \frac{1}{WH} \Sigma \left( \alpha_{i,j}^{SR} \phi_{2,2}(I_{SR}) - \alpha_{i,j}^{HR} \phi_{2,2}(I_{HR}) \right)^2 \tag{15}$$

$$L_{high-level} = \frac{1}{WHC} \Sigma \left( \phi_{5,4}(I_{SR}) - \phi_{5,4}(I_{HR}) \right)^2 \tag{16}$$

where $\phi(\cdot)$ denotes the feature maps, $W$, $H$ and $C$ are the width, height and channel of the feature maps respectively, $\alpha_{i,j}$ denotes the spatial weight, which is applied to each channel of feature maps $\phi_{2,2}(\cdot)$, and $i$ and $j$ denote the horizontal and vertical indexes of spatial weight separately. The calculation of $\alpha_{i,j}$ is summarized in Algorithm 1.

Fig. 3 shows the comparison experiment based on content loss function and weighted content loss function. As shown in Fig. 3(c), the difference between salient regions and non-salient regions in the image is extended after weighting feature maps. The feature map is got by fusing the feature maps (56×56×128) extracted from $\phi_{2,2}$, according to 1:1. Compared with content loss, weighted content loss is helpful to enhance the details of salient regions and improve SR image visual quality, as shown in Fig. 3(e).

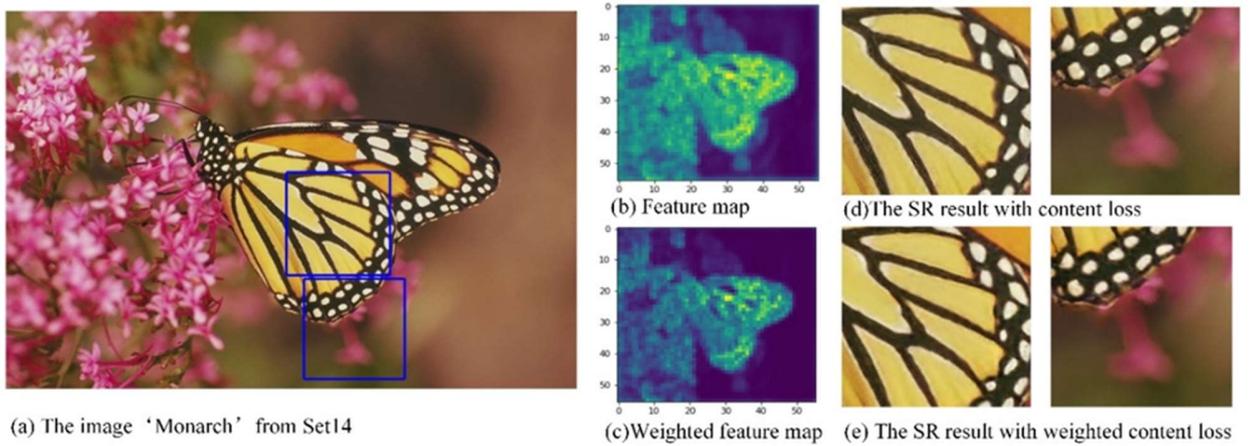

(a) The image 'Monarch' from Set14   (b) Feature map   (d)The SR result with content loss
(c)Weighted feature map   (e) The SR result with weighted content loss

**Fig. 3** Effect of weighted content loss and SR result.

---

* $\phi_{2,2}$ and $\phi_{5,4}$ denote the 2-th convolution after activation before the 2-th max-pooling layer and the 4-th convolution after activation before the 5-th max-pooling layer within the VGG19 network, respectively.

---
Algorithm 1: The calculation of spatial weight $\alpha_{i,j}$.
---

**Input**: Low-level feature maps $\chi_{kij}$. Let $\chi_{kij} \in \mathbb{R}^{C \times W \times H}$ denote the 3-dimensional feature maps extracted from the selected layer $\phi_{2,2}$. $i$, $j$ and $k$ denote horizontal, vertical and channel indexes of feature maps, respectively. $W$, $H$ and $C$ are the width, height and channel of the feature maps respectively.

1. Obtain the accumulated feature responses $f_{i,j}^C$ by summing the feature maps $\chi_{kij}$ per channel for each location. $f_{i,j}^C \in \mathbb{R}^{W \times H} (i = 1, \cdots W, j = 1, \cdots H)$.

$$f_{i,j}^C = \sum_{k=1}^{C} \chi_{k,i,j}$$

2. Normalize accumulated feature maps by $L2$ norm to calculate spatial weight $\alpha_{i,j}$.

$$\alpha_{i,j} = \frac{f_{i,j}^C}{\sqrt{\left(\sum_{i=1}^{H} \sum_{j=1}^{W} (f_{i,j}^C)^2\right)}}$$

3. Obtain the weighted feature maps $\hat{\chi}_{k,i,j}$.

$$\hat{\chi}_{k,i,j} = \chi_{k,i,j} \cdot \alpha_{i,j}$$

**Output**: Spatial weight $\alpha_{i,j}$.

---

*3.2.4  Total Loss Function*

Our perceptual loss function is a weighted sum of loss functions mentioned above, defined as follows:

$$L = L_{pixel} + 10^{-3} L_{adv} + 2 \times 10^{-4} L_{wc} \tag{17}$$

## 4  Experiments

In this section, we conduct the numerous experiments on benchmark datasets to verify the performance of the proposed methods and compare them with a series of state-of-the-art methods based on different loss function. All experiments are implemented on NVIDIA GeForce GTX 1080ti (12G memory).

*4.1 Experimental Data*

*4.1.1 Datasets*

During training, we train our SR model with a scale factor of × 4 using 800 images (1300 × 2000 pixels) from DIV2K dataset. DIV2K dataset includes lots of high-resolution RGB images with a large diversity of contents (such as flora, fauna, handmade object, people, scenery, etc.). It captures sufficient variability of natural images and provides abundant information for SR network learning. The corresponding LR images are obtained using Matlab imresize function in MATLAB R2016b.

During testing, we perform experiments on three benchmark datasets Set5[29], Set14[30] and BSD100[31]. The Set5 dataset contains 5 images: 'baby', 'bird', 'butterfly', 'head' and 'woman'. The Set14 dataset has 14 images. Some images include complex edges and textures (e.g., 'baboon', 'comic', 'face', etc.), some images include more edges than textures (e.g., 'monarch', 'barbara', etc.), others include rich textures (e.g., 'coastguard', 'zebra', 'flowers', etc.). The BSD100 dataset, which has 100 images, is built by UC Berkeley Computer Vision Group. This dataset contains different categories (such as animal, building, food, landscape, people, plant, etc.). The information contained in the above three datasets is all-encompassing. It is available to measure the robustness of the different SR methods.

*4.1.2 Compared Method*

GAN-IMC introduces morphological component adversarial loss and color adversarial loss. It is trained with a combination of pixel loss, image adversarial loss, morphological component adversarial loss, color adversarial loss and content loss. Optimized method GAN-IMCW is trained using weighted content loss instead of content loss in GAN-IMC.

We compare the proposed method with the bicubic method and several the state-of-the-art SR methods, including VDSR[13], EDSR[16], SRGAN-MSE[17], SRGAN-VGG22[17], SRGAN[17], EnhanceNet[19] and SRFeat[22]. Table 2 shows the loss functions of the compared methods. Both VDSR and EDSR trained by pixel loss belong to PSNR-driven SR methods. SRGAN-MSE, SRGAN, EnhanceNet and SRFeat trained with the combination of different loss belong to perceptual-quality driven SR methods. SRGAN-MSE trained with the combination of pixel loss and image adversarial loss is a baseline method.

**Table 2** Comparison of different perceptual loss function among the compared methods.

| Loss function | Image pixel loss | Adversarial loss | | | | Content loss | | Texture loss |
|---|---|---|---|---|---|---|---|---|
| | | Image adversarial loss | Feature adversarial loss | Morphological component adversarial loss | Color adversarial loss | Content loss | Weighted content loss | |
| VDSR | √ | | | | | | | |
| EDSR | √ | | | | | | | |
| SRGAN-MSE (baseline) | √ | √ | | | | | | |
| SRGAN | | √ | | | | √ | | |
| EnhanceNet | √ | √ | | | | √ | | √ |
| SRFeat | √ | √ | √ | | | √ | | |
| GAN-IMC (Ours) | √ | √ | | √ | √ | √ | | |
| GAN-IMCW (Ours) | √ | √ | | √ | √ | | √ | |

*4.1.3 Evaluation*

We apply PI and NIQE indicators to compare the performance of SR methods. The PSNR is also used to evaluate the image distortion in Sec 4.3. We have provided more details about the calculation of PI and NIQE in Sec 2.2. All experiment evaluation values of competing methods are from their publication and SR results of competing methods are from the author's website.

*4.1.4 Parameters setting*

The experimental parameters of our method are set as follows. We crop randomly 16 $96 \times 96$ sub-images from 800 images for each training batch. The Adam optimizer with $\beta_1 = 0.9$ is applied. The SR network is pre-trained using pixel loss function with a learning rate of $10^{-4}$ and $5 \times 10^4$ update iterations for the initialization of the generator. We alternately update the generator and discriminators with a learning rate of $10^{-4}$. After $10^5$ update iterations, the learning rate is reduced by ten times.

4.2 Experimental Results

We conduct the numerous experiments on benchmark datasets Set5, Set14 and BSD100 and show some SR results. In order to clearly compare, we amplify two times of local line in the left upper corner of the figure.

*4.2.1 Set5 dataset*

We experimented with all 5 LR test images in the Set 5 dataset. In Fig. 4-5, we show $\times 4$ SR results on images 'baby' and 'head'. Table 3 shows the PI and NIQE values of nine methods on all images in Set5.

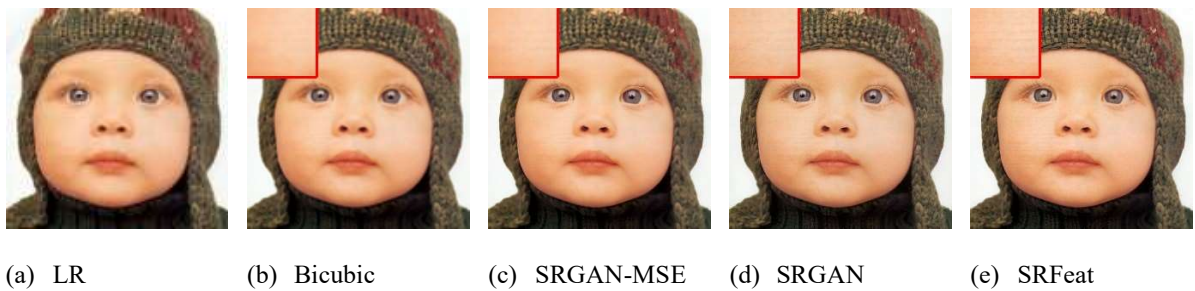

(a)  LR         (b)  Bicubic      (c)  SRGAN-MSE    (d)  SRGAN      (e)  SRFeat

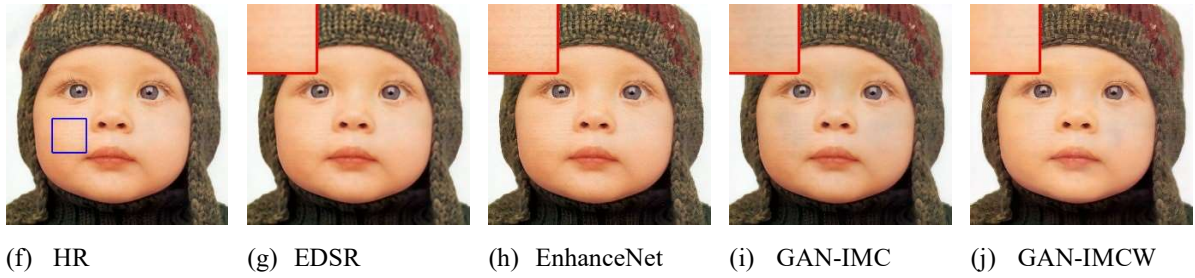

(f) HR     (g) EDSR     (h) EnhanceNet     (i) GAN-IMC     (j) GAN-IMCW

**Fig. 4** The SR results of 'baby' (upscaling factor of 4).

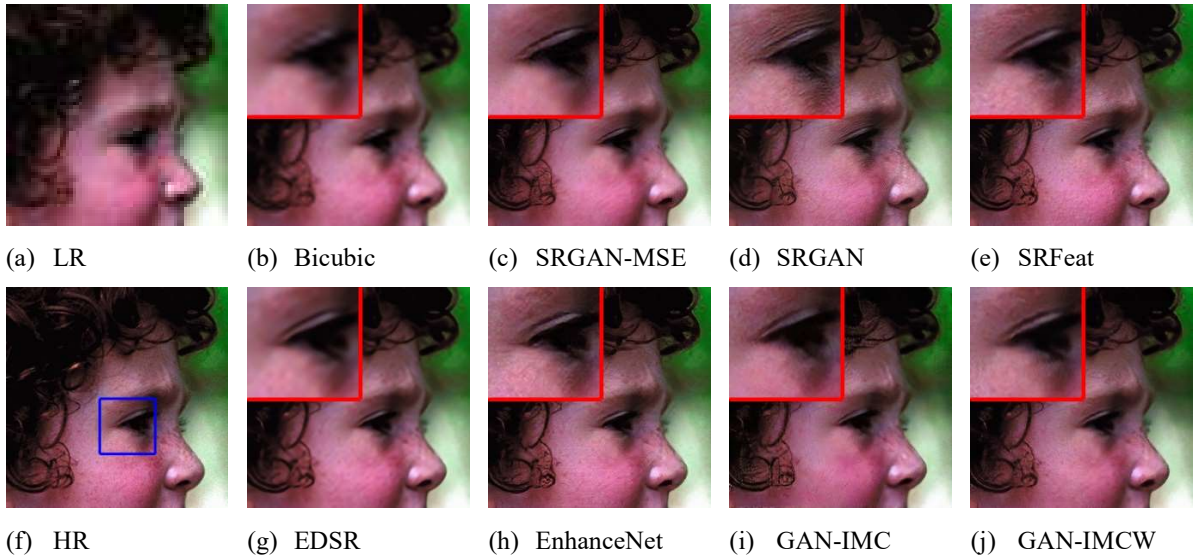

(a) LR     (b) Bicubic     (c) SRGAN-MSE     (d) SRGAN     (e) SRFeat

(f) HR     (g) EDSR     (h) EnhanceNet     (i) GAN-IMC     (j) GAN-IMCW

**Fig. 5** The SR results of 'head' (upscaling factor of 4).

**Table 3** PI and NIQE values on Set 5.

| Image | VDSR | EDSR | SRGAN-MSE (baseline) | SRGAN | EnhanceNet | SRFeat | GAN-IMC | GAN-IMCW |
|---|---|---|---|---|---|---|---|---|
| baby | 5.896/6.759 | 5.915/6.741 | 4.349/4.435 | 2.642/3.681 | 2.572/4.072 | 2.563/3.553 | 2.041/3.081 | 2.368/3.465 |
| bird | 6.331/7.457 | 6.060/6.802 | 3.563/5.010 | 3.398/4.739 | 2.613/3.750 | 3.576/4.893 | 2.743/4.050 | 3.362/4.691 |
| butterfly | 6.893/10.501 | 5.391/7.729 | 3.492/5.290 | 3.523/5.541 | 2.943/4.500 | 4.620/7.551 | 3.328/5.442 | 4.147/7.069 |
| head | 6.734/8.197 | 6.505/7.330 | 4.790/5.847 | 4.305/5.270 | 3.803/4.753 | 3.965/4.817 | 5.125/6.771 | 4.524/5.580 |
| woman | 5.630/6.128 | 5.657/6.278 | 3.028/4.402 | 2.909/4.023 | 2.699/4.247 | 2.849/4.221 | 2.920/4.713 | 2.816/4.478 |
| AVE | 6.297/7.812 | 5.906/6.976 | 3.844/4.997 | 3.355/4.650 | 2.926/4.464 | 3.515/5.007 | 3.231/4.811 | 3.443/5.056 |

*4.2.2 Set14 dataset*

We experimented with all 14 LR test images in the Set14 dataset. Their × **4** SR results on images 'coastguard', 'flowers' and 'monarch' are shown in Fig. 6-8. Table 4 shows the PI and NIQE values of nine methods on all images in Set14.

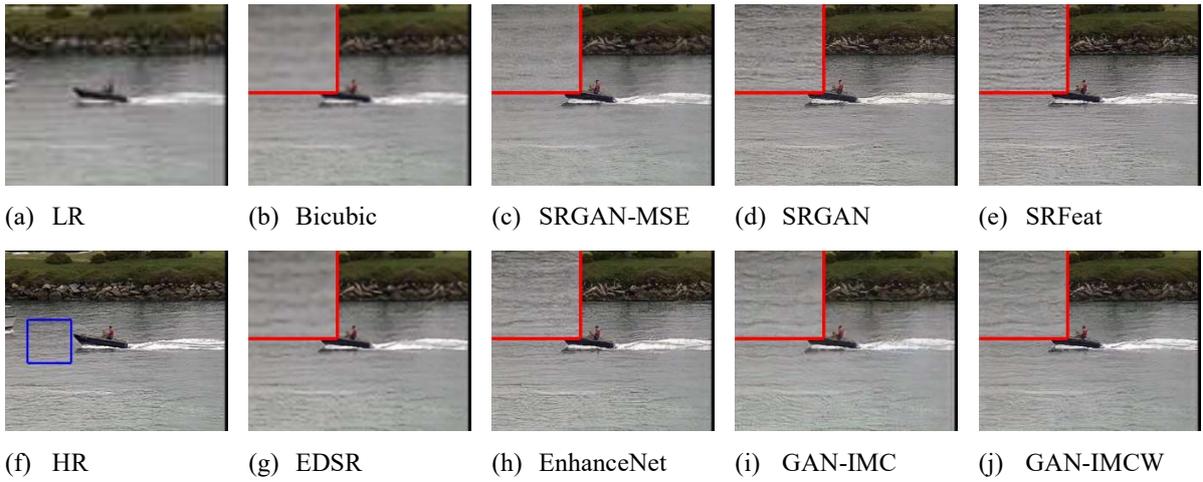

**Fig. 6** The SR results of 'coastguard' (upscaling factor of 4).

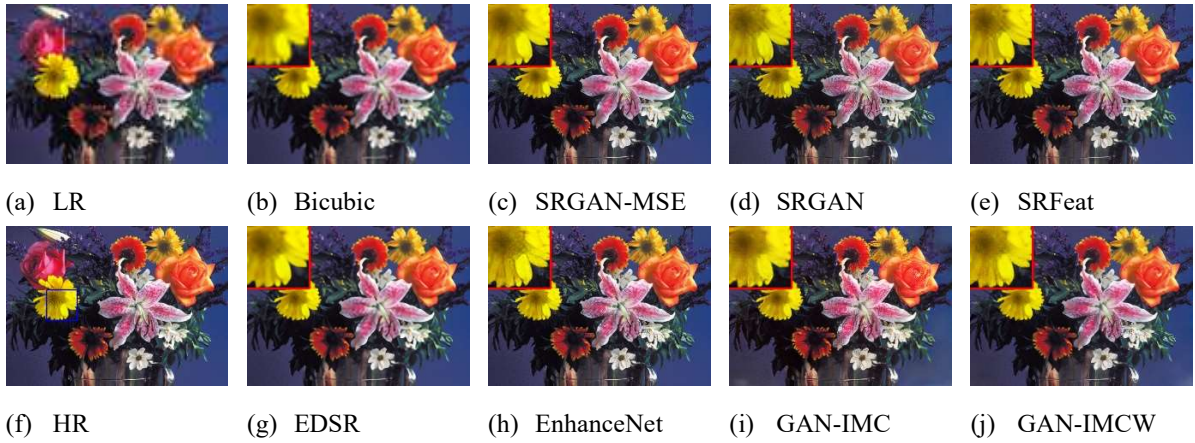

**Fig. 7** The SR results of 'flowers' (upscaling factor of 4).

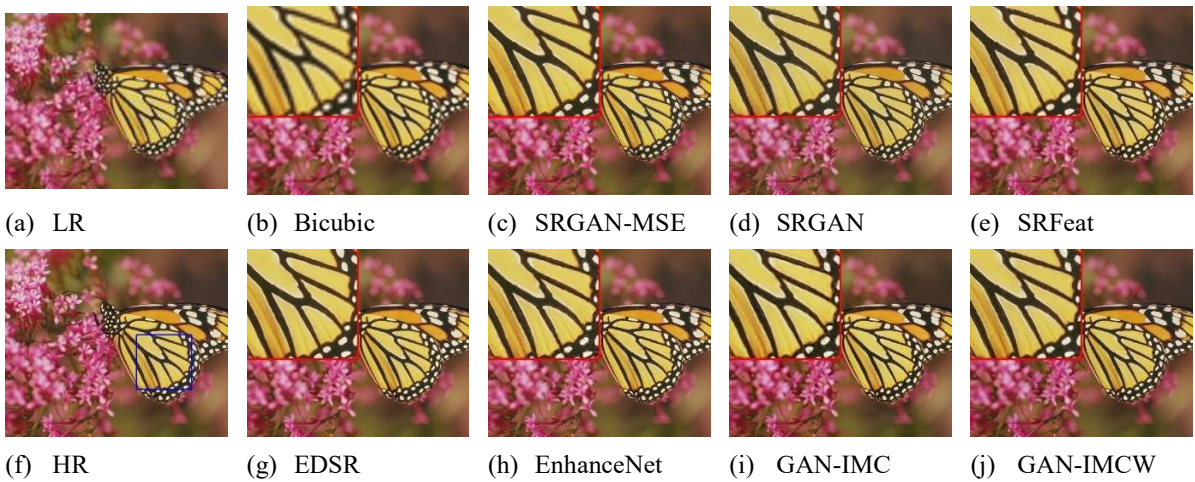

**Fig. 8** The SR results of 'monarch' (upscaling factor of 4).

*4.2.3 BSD100 dataset*

We experimented with all 100 LR test images in the BSD100 dataset. Their ×4 SR results on images '14037' and '106024' are shown in Fig. 9-10. Table 5 shows the PI and NIQE values of nine methods on some images in BSD100.

Table 4 PI and NIQE values on Set 14.

| Image | VDSR | EDSR | SRGAN-MSE (baseline) | SRGAN | EnhanceNet | SRFeat | GAN-IMC | GAN-IMCW |
|---|---|---|---|---|---|---|---|---|
| baboon | 5.636/6.896 | 4.647/4.887 | 1.710/2.363 | 2.616/4.021 | 2.666/4.179 | 1.887/2.667 | 1.890/2.624 | 1.947/2.753 |
| barbara | 5.125/5.889 | 5.282/5.991 | 1.970/2.813 | 2.057/2.779 | 2.100/2.939 | 2.082/2.893 | 2.099/3.021 | 1.934/2.833 |
| bridge | 5.590/6.575 | 5.432/6.171 | 3.014/3.432 | 1.807/2.548 | 2.459/3.899 | 1.821/2.638 | 1.974/2.914 | 2.179/3.334 |
| coastguard | 5.668/6.05 | 7.450/9.923 | 2.913/4.351 | 3.513/5.853 | 3.265/5.364 | 3.211/5.185 | 2.806/4.108 | 2.732/4.109 |
| comic | 5.898/6.613 | 3.981/5.551 | 2.424/3.806 | 2.325/3.622 | 2.403/3.731 | 2.016/2.972 | 2.126/3.230 | 1.877/2.730 |
| face | 6.929/8.196 | 7.062/7.929 | 5.037/6.285 | 4.592/5.596 | 3.732/4.537 | 4.544/5.614 | 5.190/6.741 | 4.516/5.153 |
| flowers | 4.844/5.760 | 3.605/5.142 | 2.007/3.074 | 2.010/3.067 | 2.363/3.744 | 2.051/3.074 | 2.109/3.115 | 2.088/3.249 |
| foreman | 6.307/6.974 | 6.320/7.653 | 4.114/6.184 | 3.443/4.443 | 4.157/5.435 | 4.054/5.105 | 3.588/4.718 | 3.460/4.638 |
| lenna | 5.184/6.193 | 4.916/6.014 | 2.556/3.732 | 2.707/4.314 | 2.574/3.989 | 2.472/3.689 | 2.893/4.295 | 2.764/3.999 |
| man | 4.930/5.799 | 5.053/5.579 | 2.572/3.982 | 1.722/2.458 | 2.019/3.071 | 1.911/2.810 | 2.058/2.979 | 2.063/3.112 |
| monarch | 6.009/5.901 | 5.667/5.710 | 3.847/3.678 | 3.416/3.756 | 3.553/3.685 | 3.881/4.079 | 3.332/3.346 | 3.939/4.298 |
| pepper | 5.759/6.444 | 5.688/6.403 | 3.292/4.657 | 3.169/4.424 | 3.252/4.514 | 3.620/4.877 | 3.467/4.481 | 4.121/4.606 |
| ppt3 | 5.729/6.550 | 5.393/6.414 | 3.720/4.194 | 3.982/4.495 | 3.892/4.103 | 4.387/5.160 | 3.584/3.819 | 3.737/3.968 |
| zebra | 6.133/6.702 | 6.696/6.713 | 3.722/3.524 | 2.985/2.869 | 3.813/4.624 | 3.464/3.547 | 2.908/2.757 | 2.989/2.999 |
| AVE | 5.696/6.467 | 5.514/6.434 | 3.064/4.005 | 2.882/3.875 | 3.018/4.130 | 2.957/3.879 | 2.859/3.725 | 2.881/3.698 |

Table 5 PI and NIQE values on BSD100.

| Image | VDSR | EDSR | SRGAN-MSE (baseline) | SRGAN | EnhanceNet | SFeat | GAN-IMC | GAN-IMCW |
|---|---|---|---|---|---|---|---|---|
| '101085' | 5.12/5.744 | 4.941/5.633 | 2.754/4.547 | 2.031/2.897 | 2.364/3.691 | 1.960/2.796 | 1.846/2.777 | 1.810/2.614 |
| '101087' | 4.937/6.486 | 5.248/7.013 | 2.698/4.345 | 2.334/3.667 | 2.648/4.279 | 2.229/3.447 | 2.247/3.485 | 2.182/3.403 |
| '102061' | 5.596/6.531 | 5.186/6.057 | 2.435/3.819 | 1.928/2.891 | 2.058/3.211 | 2.036/3.159 | 2.049/3.168 | 1.970/2.934 |
| '103070' | 6.124/6.451 | 6.264/6.508 | 3.167/4.195 | 1.829/2.364 | 2.538/4.134 | 2.530/3.253 | 2.344/3.221 | 2.282/3.035 |
| '105025' | 5.470/6.411 | 5.519/6.396 | 2.243/3.420 | 1.699/2.368 | 2.749/4.403 | 1.959/2.787 | 2.256/3.367 | 1.988/2.948 |
| '106024' | 7.214/7.403 | 7.403/7.796 | 3.903/5.385 | 3.197/3.956 | 4.040/5.395 | 3.548/4.267 | 3.855/4.241 | 3.253/3.874 |
| ⋮ | ⋮ | ⋮ | ⋮ | ⋮ | ⋮ | ⋮ | ⋮ | ⋮ |
| '97033' | 5.005/5.755 | 4.501/5.340 | 2.583/4.184 | 1.936/2.912 | 2.623/4.295 | 1.956/2.952 | 1.970/2.946 | 2.088/3.203 |
| AVE | 5.700/6.677 | 5.559/6.583 | 2.802/4.032 | 2.397/3.407 | 2.908/4.525 | 2.520/3.623 | 2.488/3.712 | 2.330/3.361 |

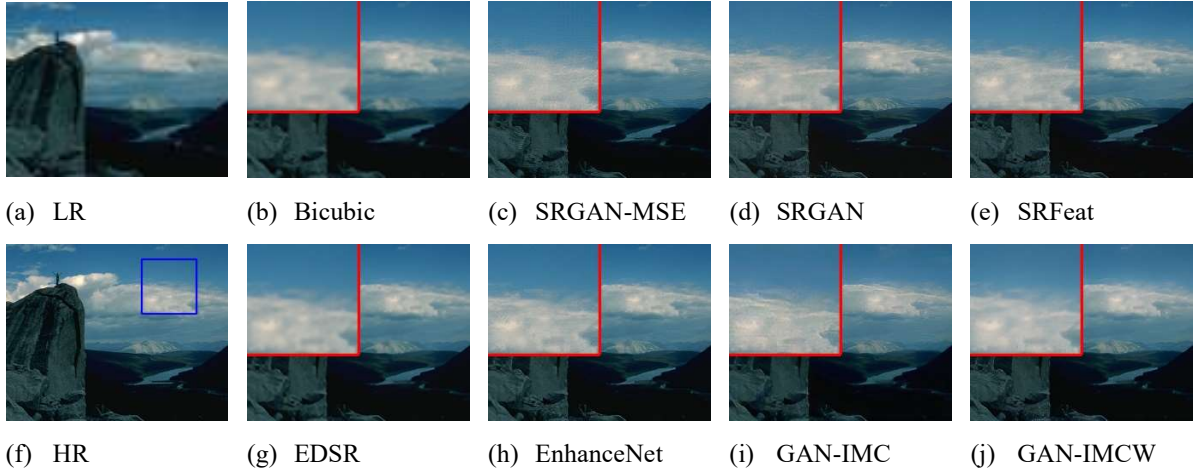

**Fig. 9** The SR results of '14037' (upscaling factor of 4).

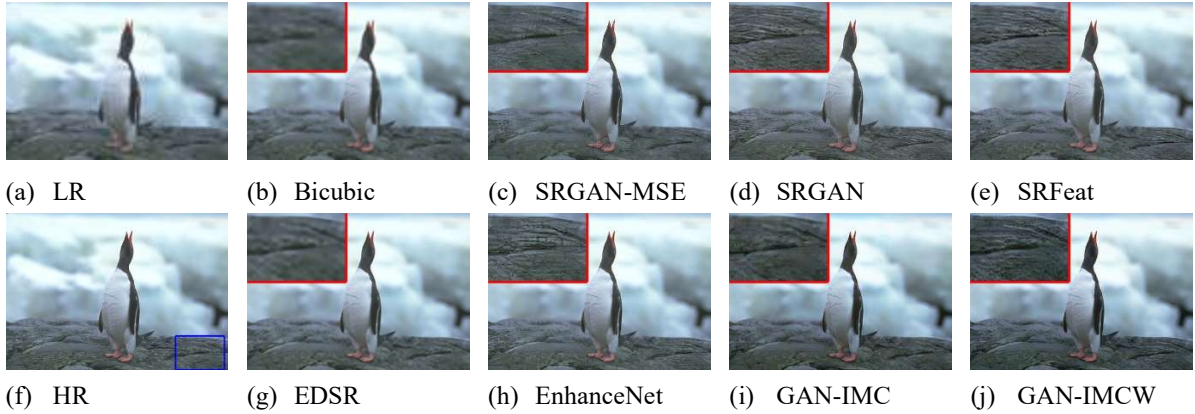

**Fig. 10** The SR results of '106024' (upscaling factor of 4).

**Table 6** The average PSNR values (dB) on three datasets.

| Dataset | bicubic | VDSR | EDSR | SRGAN-MSE (baseline) | SRGAN | EnhanceNet | SRFeat | GAN-IMC | GAN-IMCW |
|---|---|---|---|---|---|---|---|---|---|
| Set5 | 28.418 | 31.349 | 32.630 | 30.666 | 29.410 | 28.564 | 30.054 | 27.886 | 28.751 |
| Set14 | 26.090 | 28.015 | 28.953 | 27.006 | 26.114 | 25.770 | 26.722 | 25.674 | 25.960 |
| BSD100 | 25.957 | 27.287 | 27.796 | 25.981 | 25.176 | 24.930 | 25.693 | 24.201 | 24.501 |

## 4.3 Evaluation of Experiment Results

**Visual Quality.** By comparing and analyzing the experimental results of Fig. 4-10, our methods have the following conclusions: (i) The large-scale structural edges are sharp and natural. More abundant and realistic texture details are generated while avoiding over-fake textures and less texture details in SR results of GAN-IMC and GAN-IMCW. We observed that the edges of the eye, flowers and butterfly are significantly sharp from the highlighted window of Fig. 5, 7 and 8.

The highlighted window of Fig. 4, 5, 6, 9 and 10 display that the textured areas are well recovered without introducing additional visible grids and texture details are not weakened and lost. (ii) Local color transition is natural and smooth. As shown in the highlighted window of Fig. 5, 7, 8 and 9, the color transition is comfortable. It can be seen in Fig. 5 that the eyelid part of the girl generated by GAN-IMCW is highly similar to the corresponding position of the HR image. (iii) The recovery of feature-rich regions is remarkable. As shown in the highlighted window of Fig. 7-8, GAN-IMCW expands the difference between salient regions (e.g. flowers, butterfly) and backgrounds to enhance visual quality of flowers and the butterfly, and accurately recovers the details of salient regions. (iv) GAN-IMCW shows better robustness on three test datasets containing a large number of different types of images. It makes full use of edge, texture and color features of images and highlights feature-rich regions to produce visually pleasant SR results.

In summary, compared with the competing method, GAN-IMC recovers sharp large-scale structural edges and realistic texture details, and SR results have smooth color transition. Furthermore, GAN-IMCW significantly improves the visual attention of feature-rich regions in SR results. GAN-IMCW has achieved better performance in terms of both effectiveness and robustness, which shows good agreement with the quantitative evaluation results in Table 3, 4 and 5.

**PI and NIQE.** The PI, NIQE, average PI and average NIQE values of the competing methods on all images in Set5, Set14 and BSD100 are shown in Table 3, 4 and 5. GAN-IMC and GAN-IMCW achieve much better PI and NIQE index than SRGAN-MSE (baseline) on three datasets. For the PI, GAN-IMCW is better than SRGAN and EnhanceNet on Set14 and BSD100, and better than SRFeat on Set5, Set14 and BSD100. Our average gain on PI for BSD100 is 0.067, 0.578 and 0.19 less than the value of SRGAN, EnhanceNet and SRFeat separately. For the NIQE, GAN-IMCW is better than EnhanceNet on Set14 and BSD100, and better than both SRFeat and SRGAN on Set5, Set14 and BSD100. The results demonstrate that GAN-IMCW has good robustness for different kinds of test images and significantly improves the perceptual quality of SR results.

**PSNR.** The average PSNR values of the different methods at ×4 SR on three benchmark datasets are shown in Table 6. Compared with the competing method, the PSNR values of the proposed method do not increase but decrease, which also proves that the proposed method can effectively improve image perceptual quality, as described in Sec. 2.1.

## 5 Conclusion

In this paper, we designed a novel SR network framework GAN-IMC that includes a generator and multi-feature discriminators. The data distribution of the image and component features of the image, including color, edge and texture are learned during adversarial learning between generator network and multi-feature discriminator network. Moreover, the optimized method GAN-IMCW improves the visual quality of feature-rich regions in images by using weighted content loss. A large number of experimental results indicated the superiority of GAN-IMCW over the other competing methods. It not only achieves competitive PI and NIQE values but also improves more pleasant visual quality in terms of image, edge, texture, color and feature-rich regions.

*Acknowledgments*

This work was supported by the key project of Natural Science Foundation of Shaanxi Province (Grant Nos. 2018JZ6007).